%% file: main.tex
\documentclass[10pt,twocolumn,letterpaper]{article}

\usepackage[pagenumbers]{cvpr} 

\usepackage{graphicx}
\usepackage{amsmath}
\usepackage{amssymb}
\usepackage{booktabs}
\usepackage{algorithm}
\usepackage[noend]{algorithmic}
\usepackage{balance}

\DeclareFontFamily{OT1}{pzc}{}
\DeclareFontShape{OT1}{pzc}{m}{it}{<-> s * [1.10] pzcmi7t}{}
\DeclareMathAlphabet{\mathpzc}{OT1}{pzc}{m}{it}

\usepackage[pagebackref,breaklinks,colorlinks]{hyperref}
\usepackage[capitalize]{cleveref}
\crefname{section}{Sec.}{Secs.}
\Crefname{section}{Section}{Sections}
\Crefname{table}{Table}{Tables}
\crefname{table}{Tab.}{Tabs.}

\newcommand{\mmL}{F}

\def\cvprPaperID{9613} 
\def\confName{CVPR}
\def\confYear{2022}

\input{preamble}
\begin{document}
\input{sec/0_metadata}
\maketitle
\input{sec/0_abstract}
\input{sec/1_introduction}
\input{sec/2_related}

\input{sec/3_method}
\input{sec/4_results}

\input{sec/5_conclusions}

{
    \small
    \balance
    \bibliographystyle{ieee_fullname}
    \bibliography{macros,main}
}

\input{sec/X_supplementary}


\end{document}

%% file: preamble.tex
\usepackage{overpic}
\usepackage{enumitem} 
\usepackage{overpic} 
\usepackage{color}

\definecolor{turquoise}{cmyk}{0.65,0,0.1,0.3}
\definecolor{purple}{rgb}{0.65,0,0.65}
\definecolor{dark_green}{rgb}{0, 0.5, 0}
\definecolor{orange}{rgb}{0.8, 0.6, 0.2}
\definecolor{red}{rgb}{0.8, 0.2, 0.2}
\definecolor{darkred}{rgb}{0.6, 0.1, 0.05}
\definecolor{blueish}{rgb}{0.0, 0.3, .6}
\definecolor{light_gray}{rgb}{0.7, 0.7, .7}
\definecolor{pink}{rgb}{1, 0, 1}
\definecolor{greyblue}{rgb}{0.25, 0.25, 1}

\usepackage{blindtext}

\renewcommand{\paragraph}[1]{\vspace{1em}\noindent\textbf{#1}.}

%% file: sec/0_metadata.tex
\title{SPIDER: Searching Personalized Neural Architecture for Federated Learning}

\author{
    Erum Mushtaq \textsuperscript{\rm 1}, Chaoyang He \textsuperscript{\rm 1}, Jie Ding \textsuperscript{\rm 2}, Salman Avestimehr \textsuperscript{\rm 1}\\
}


\author{Erum Mushtaq\\
University of Southern California\\
{\tt\small emushtaq@usc.edu} \and Chaoyang He\\
University of Southern California\\
{\tt\small chaoyang.he@usc.edu} \and Jie Ding \\ University of Minnesota\\
{\tt\small dingj@umn.edu} \and Salman Avestimehr \\University of Southern California\\
{\tt\small avestime@usc.edu}
}

%% file: sec/0_abstract.tex
\begin{abstract}


Federated learning (FL) is an efficient learning framework that assists distributed machine learning when data cannot be shared with a centralized server due to privacy and regulatory restrictions. Recent advancements in FL use predefined architecture-based learning for all the clients. However, given that clients' data are invisible to the server and data distributions are non-identical across clients, a predefined architecture discovered in a centralized setting may not be an optimal solution for all the clients in FL. Motivated by this challenge, in this work, we introduce SPIDER, an algorithmic framework that aims to \underline{\textbf{S}}earch \underline{\textbf{P}}ersonal\underline{\textbf{I}}zed neural architecture for fe\underline{\textbf{DER}}ated learning. SPIDER is designed based on two unique features: (1) alternately optimizing one architecture-homogeneous global model (Supernet) in a generic FL manner and one architecture-heterogeneous local model that is connected to the global model by weight sharing-based regularization (2) achieving architecture-heterogeneous local model by a novel  neural architecture search (NAS) method that can select optimal subnet progressively using operation-level perturbation on the accuracy value as the criterion. Experimental results demonstrate that SPIDER outperforms other state-of-the-art personalization methods, and the searched personalized architectures are more inference efficient.

\end{abstract}

%% file: sec/1_introduction.tex
\section{Introduction}
\label{sec:intro}






Federated Learning (FL) is a promising decentralized machine learning framework that facilitates data privacy and low communication costs. It has been extensively explored in various machine learning domains such as computer vision, natural language processing, and data mining. Despite many benefits of FL, one major challenge involved in FL is data heterogeneity, meaning that the data distributions across clients are not identically or independently (non-I.I.D) distributed. The non-I.I.D distributions result in the varying performance of a globally learned model across different clients. In addition to data heterogeneity, data invisibility is another challenge in FL. Since clients' private data remain invisible to the server, from the server's perspective, it is unclear how to select a pre-defined architecture from a pool of all available candidates. In practice, it may require extensive experiments and hyper-parameter tuning over different architectures, a procedure that can be prohibitively expensive.

To address the data-heterogeneity challenge, variants of the standard FedAvg have been proposed to train a global model, including the \texttt{FedProx} \cite{li2018federated}, \texttt{FedOPT} \cite{reddi2020adaptive}, and \texttt{FedNova} \cite{wang2020tackling}. In addition to training of a global model, frameworks that focus on training personalized models have also gained a lot of popularity. The \texttt{Ditto} \cite{li2021ditto}, \texttt{PerFedAvg} \cite{fallah2020personalized}, and \texttt{pFedMe} \cite{dinh2020personalized} are some of the recent works that have shown promising results to obtain improved performance across clients. However, all these works exploit pre-defined architectures and operate at the optimization layer. Consequently, in addition to their inherent hyper-parameters tuning, these personalization frameworks often encounter the data-invisibility challenge that one has to select a suitable model architecture involving a lot of hyper-parameter tuning. 

In this work, we adopt a different and complementary technique to address the data heterogeneity challenge for FL. We introduce SPIDER, an algorithmic framework that aims to \underline{\textbf{S}}earch \underline{\textbf{P}}ersonal\underline{\textbf{I}}zed neural architecture for fe\underline{\textbf{DER}}ated learning. Recall that in a centralized setting, the neural architecture search (NAS) aims to search for optimal architecture to address system design challenges such as lower latency \cite{wu2019fbnet}, lesser memory cost \cite{li2021hw}, and smaller energy consumption \cite{yang2020co}. For architecture search, there are three well known methods explored in literature, gradient-based~\cite{liu2018darts}, evolutionary search~\cite{liu2021survey}, and reinforcement learning~\cite{jaafra2019reinforcement}. Out of these, gradient-based methods are generally considered more efficient because of their ability to yield higher performance in comparatively lesser time~\cite{santra2021gradient}. 

To achieve personalization at the architecture level in FL,
we propose a unified framework, SPIDER. This framework essentially deploys two models, local and global models, on each client. Initially, both models use the DARTS search space-based Supernet~\cite{liu2018darts}, an over-parameterized architecture. In the proposed framework, the global model is shared with the server for the FL updates and, therefore, stays the same in the architecture design. On the other hand, the local model stays completely private and performs personalized architecture search, therefore, gets updated. To search for the personalized child model, SPIDER deploys SPIDER-Searcher on each client's local model. The SPIDER-Searcher is built upon a well-known gradient-based NAS method, named perturbation-based NAS \cite{wang2021rethinking}. The main objective of the SPIDER framework is to allow each client to search and optimize their local models while benefiting from the global model. To achieve this goal, we propose an alternating bi-level optimization-based SPIDER Trainer that trains local and global models in an alternate fashion. However, the main challenge here is the optimization of an evolving local model architecture while benefiting from a fixed global architecture. To address this challenge, SPIDER Trainer performs weight sharing-based regularization that regularizes the common connections between the global model's Supernet and the local model's child model. This aids clients in searching and training heterogeneous architectures tailored for their local data distributions. In a nutshell, this approach not only yields architecture personalization in FL but also facilitates model privacy (in the sense that the derived child local model is not shared with the server at all).


To evaluate the performance of the proposed algorithm, we consider a cross-silo FL setting and use Dirichlet distribution to create non-I.I.D data distribution across clients. For evaluation, we report test accuracy at each client on the 20\% of training data kept as test data for each client. We show that the architecture personalization yields better results than state-of-the-art personalization algorithms based solely on the optimization layer, such as Ditto \cite{li2021ditto}, perFedAvg \cite{fallah2020personalized}, and local adaptation~\cite{Cheng2021FinetuningIF}.

To summarize, the following are the key contributions of our work.

\noindent $\bullet$ We propose and formulate a personalized neural architecture search framework for FL named SPIDER, from a perspective complementary to the state-of-the-arts to address data heterogeneity challenges in FL.

\noindent $\bullet$ SPIDER is designed based on two unique features: (1) maintaining two models at each client, one to communicate with the server and the other to perform a local progressive search, and (2) operating local search and training at each client by an alternating bilevel optimization and weight sharing-based regularization along the FL updates.

\noindent $\bullet$ We run extensive experiments to demonstrate the benefit of SPIDER compared with state-of-the-art personalized FL approaches such as Ditto~\cite{li2021ditto}, perFedAvg~\cite{fallah2020personalized} and Local Adaptation~\cite{Cheng2021FinetuningIF}. In particular, on the CIFAR10 dataset with heterogeneous distribution, we demonstrate an increase of the average local accuracy by 2.8\%, 1.7\%, and 5.5\%, over Ditto, PerFedAvg, and Local Adaption, respectively.

\noindent $\bullet$ We also demonstrate that SPIDER learns smaller personalized architectures of average size around 14MB, which is three times smaller than a pre-defined architecture Resnet18 (of size 44MB). 

%% file: sec/2_related.tex
\input{fig/overview}

\section{Related works}
\paragraph{Heterogeneous Neural Architecture for FL} Heterogeneous neural architecture is one way to personalize the model in FL.
For personalization, the primal-dual framework \cite{smith2017federated}, clustering \cite{sattler2020clustered}, fine-tuning with transfer learning \cite{yu2020salvaging}, meta-learning \cite{fallah2020personalized}, regularization-based method \cite{hanzely2020federated,li2021ditto} are among the popular methods explored in the FL literature. 
Although these techniques achieve improved personalized performance, all of them use a pre-defined architecture for each client. 
HeteroFL~\cite{diao2020heterofl} is a recent work that accomplishes the aggregation of heterogeneous models by assigning sub-parts of the global model based on their computation budget and aggregating the parameters common between different clients. Another work \cite{lin2020ensemble} accomplishes this task by forming clusters of clients of the same model and allowing for heterogeneous models across clusters. On the server-side, the aggregation is based on cluster-wise aggregation followed by a knowledge distillation from the aggregated models into the global model. Given data invisibility in FL, deciding which pre-defined architecture would work for which client is a challenging task and requires exploration. As such, our proposed method aims to achieve personalized architecture automatically. 

\paragraph{Neural Architecture Search for FL} 
Neural Architecture Search (NAS) has gained momentum in recent literature to search for a global model in a federated setting. FedNAS \cite{he2020fednas} explores the compatibility of MileNAS solver with Fed averaging algorithm to search for a global model. Direct Federated NAS \cite{hu2020dsnas} is another work in this direction that explores the compatibility of a one-shot NAS method, DSNAS \cite{hu2020dsnas}, with Fed averaging algorithm with the same application, in search of a global model. \cite{zhu2021real} uses evolutionary NAS to design a master (global) model. \cite{singh2020differentially} explores the concept of differential privacy using DARTs solver \cite{liu2018darts} to explore the trade-off between accuracy and privacy of a global model. \cite{xu2020federated} starts with a pre-trained handcraft model and continues pruning the model until it satisfies the efficiency budget. Where all these models search for a unified global model, a key distinction of our work with these works is that we aim to search for a personalized model for each client.

%% file: sec/3_method.tex
\section{Preliminaries, Motivation, and Design Goals}
\label{preliminaries}

In this section, we introduce the state-of-the-art methods for personalized federated learning, discuss the motivation for personalizing model architectures, and summarize our design goals.

\paragraph{Personalized Federated Learning}
A natural formulation of FL is to assume that among $K$ distinct clients, each client $k$ has its own distribution $P_i$, draws data observations (samples) from $P_i$, and aims to solve a supervised learning task (e.g., image classification) by optimizing a global model $w$ with other clients collaboratively. At a high-level abstraction, the  optimization objective is then defined as:

\begin{align}
\min_{w^{*}}G\left(F_1(w),. . .,F_{K}(w)\right),
\end{align}
where $F_{k}(w)$ measures the performance of the model global $w$ on the private dataset at client $k$ (local objective), and $G$ is the global model aggregation function that aggregates each client's local objectives. For example, for FedAvg, $G(.)$ would be weighted aggregation of the local objectives, $\sum_{k=1}^{K} p_k F_k(w)$, where $\sum_{k=1}^{K} p_k = 1$.  

However, as distributions across individual clients are typically heterogeneous (i.e., non-I.I.D.), there is a growing line of research that advocates to reformulate FL as a personalization framework, dubbed as personalized FL (PFL). In PFL, the objective is redirected to find a personalized model $v_k$ for device $k$ that performs well on the local data distribution:
\begin{align}
\min_{v_1^{*},...,v_K^{*}}\left(F_1(v_1),. . .,F_{K}(v_K)\right),
\end{align}

To solve this challenging problem, various PFL methods are proposed, including FedAvg with local adaptation (Local-FL) \cite{Cheng2021FinetuningIF,Yu2020SalvagingFL,Wang2019FederatedEO}, MAML-based PFL (MAML-FL) \cite{Fallah2020PersonalizedFL,Jiang2019ImprovingFL},  clustered FL (CFL) \cite{Ghosh2020AnEF,Sattler2021ClusteredFL}, personalized layer-based FL (PL-FL) \cite{Liang2020ThinkLA}, federated multitask learning (FMTL) \cite{Smith2017FederatedML}, and knowledge distillation (KD) \cite{Lin2020EnsembleDF,He2020GroupKT}.

\paragraph{Motivation for Neural Architecture Personalization} 
Distinct from these existing works on PFL, we propose a new approach to instead personalize model architecture for each client. We are motivated by three key potential benefits. First, the searched architecture at each client is expected to fit its own distinct distribution, which has the potential to provide a substantial improvement over the existing PFL baselines that only personalize model weights.
Second, a personalized architecture search can result in a more compressed model for each client that will reduce inference latency and efficiency.
Third, a personalized architecture search allows the clients to even keep their local model architectures private in a sense the server and other clients neither know the architecture nor the weights of that architecture. This further enhances the privacy guarantees of FL and is helpful in business cases that each client hopes to also protect its model architecture. 


\paragraph{Design Goals} Our goal is to enable personalized neural architecture search for all clients in FL. In this context, the limitation of existing personalized FL methods is obvious: Local-FL and MAML-FL need every client to have the same architecture to perform local adaptations; In CFL, the clustering step requires all clients to share a homogeneous model architecture; PL-FL can only obtain heterogeneous architectures for a small portion of personalized layers, but it does not provide an architecture-agnostic method to determine the boundary of personalized layers in an automated mechanism; FMTL is a regularization-based method which cannot perform regularization when architectures are heterogeneous across clients; KD has an unrealistic assumption that the server has enough public dataset as the auxiliary data for knowledge distillation. In addition, an ideal FL framework for deployment is the one that can jointly optimize the inference latency/efficiency during training. However, after federated training, none of these PFL clearly specify the method for efficient inference. An additional model compression procedure (e.g., pruning or KD-based) may be required, but it is impractical to perform a remote model compression client by client.

To circumvent these limitations, our goal is to design an architecture-personalized FL framework with the following requirements: 
\begin{itemize}
    \item \textbf{R1}: \textit{allowing heterogeneous architectures for all clients}, which can capture fine-grained data heterogeneity;
    \item \textbf{R2}: \textit{searching and personalizing the entire architecture space}, to avoid the heuristic search for the boundary of personalized layers;
    \item \textbf{R3}: \textit{requiring no auxiliary data at the client- or server-side (unlike knowledge distillation-based PFL)};
    \item \textbf{R4}: \textit{resulting in models with practical inference latency and efficiency}, to avoid the need for an additional model compression procedure at each client.
\end{itemize}

We now introduce SPIDER that meets the above requirements in a unified framework.

\section{Methodology: SPIDER}

\subsection{Overview}

\begin{figure*}
\begin{center}
  \includegraphics[width=1.0\textwidth]{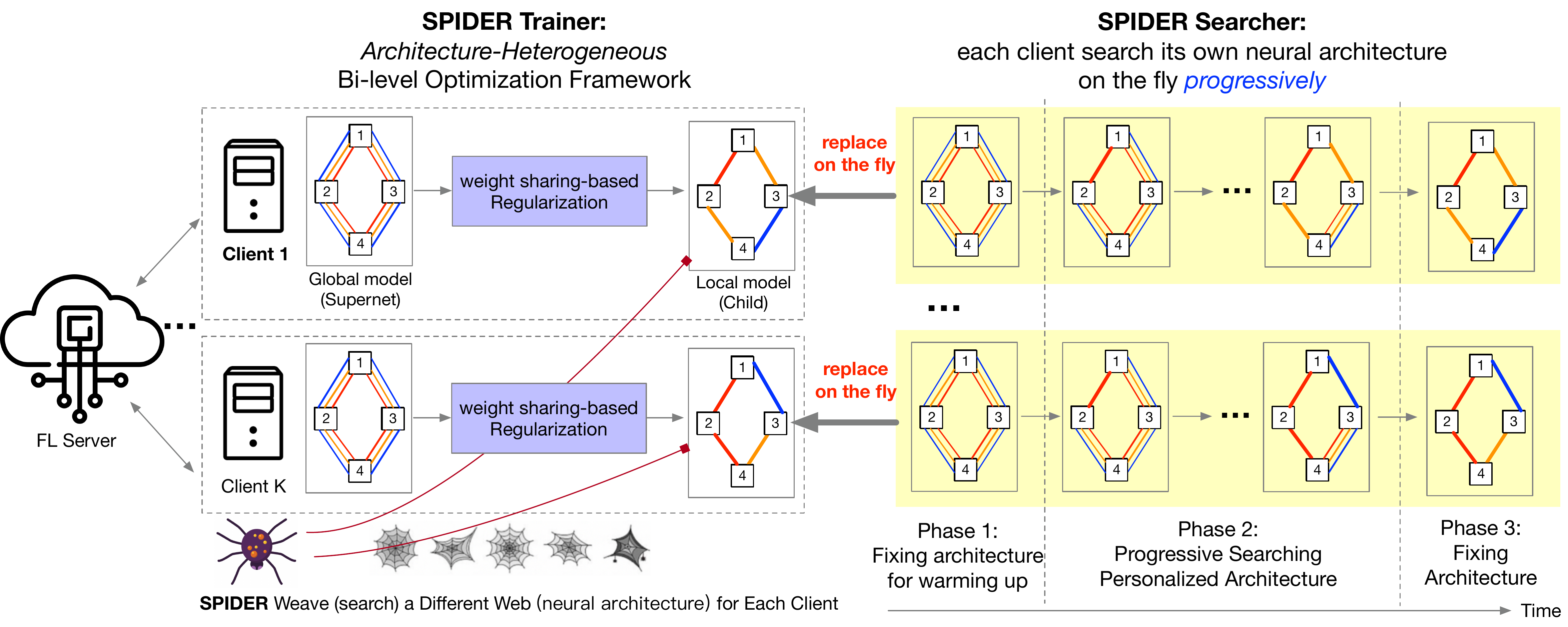}
  \caption{Illustration of SPIDER framework. SPIDER is weaving (searching) a different web (neural architecture) for each client.
 }
  \label{fig_overview}
\end{center}
\vspace{-0.8cm}
\end{figure*}

The overall framework of SPIDER is illustrated in Figure~\ref{fig_overview}. Essentially, each client maintains two models in this framework: one architecture-homogeneous global model for collaborative training with other clients, and one architecture-heterogeneous local model that initially shares the same super architecture space as the global model. 
At a high-level, SPIDER is formulated as an \textbf{architecture-personalized bi-level optimization} problem (Section \ref{sec_bilevel}) and proposes the solver as the orchestration of \textbf{SPIDER Trainer} (Section \ref{SPIDER_Train}) and \textbf{SPIDER-Searcher} (Section \ref{sec_progressive_nas}). \textbf{SPIDER Trainer} is an architecture-personalized training framework that can collaboratively train heterogeneous neural architectures across clients.
To allow federated training on the expected heterogeneous local architectures, it enables regularization between an arbitrary personalized architecture and the global model via weight sharing. With this support, \textbf{SPIDER-Searcher} is designed to \textit{dynamically adjust the architecture of each client's local model on the way}. To search a personalized architecture for the local data distribution of each client, SPIDER-Searcher is built on a novel neural architecture search (NAS) method that searches optimal local Subnet progressively using operation-level perturbation on the accuracy value as the criterion. Overall, each client's local model goes through three phases (also shown in Figure~\ref{fig_overview}): pre-training to warm up the initial local model, progressive neural architecture search, and final training of the searched architecture-personalized model.

SPIDER can meet the design goals \textbf{R1}-\textbf{R4} introduced in Section \ref{preliminaries} because 1) each client performs independent architecture personalization with its own private data (\textbf{R1}), 2) the search space is not restricted to a portion of the model (\textbf{R2}), 3) no auxiliary data is used to assist the search and train process (\textbf{R3}), and 4) progressive operation selection gradually reduces the number of candidate architectures, leading to a sparse and efficient model for inference (\textbf{R4}).

\subsection{SPIDER Formulation: Architecture-personalized Bi-level Optimization}
\label{sec_bilevel}
SPIDER aims to personalize (weave) a different neural architecture (web) for each client.
To generate heterogeneous architectures across clients, we use two models, a local model (${a}_k$) and a global model (Supernet $\mathcal{A}$) at each client, and formulate SPIDER as an architecture-personalized bi-level optimization problem for each client $k \in [K]$: 

\begin{align}
\label{eq:inner} 
&\min_{v_{k}, a_{k} \subseteq \mathcal{A}}\quad F_{k}(v_{k}, a_{k}; w^{*}, \mathcal{A})\\
\label{eq:outer} &\text{s.t.}\quad w^{*}\in\arg\min_{w}G\left(F_1(w, \mathcal{A}),. . .,F_{K}(w, \mathcal{A})\right),
\end{align}
where $F_{k}$ is the local objective of  the client $k$; $w$, $v_k$, and $a_{k}$ are all learnable parameters; $w$ denotes the parameter of the global model architecture $\mathcal{A}$, $v_{k}$ is the weight parameter of the local personalized architecture $a_{k}$ of the client $k$. Here, $a_{k}$ is a child neural architecture of a Supernet $\mathcal{A}$, denoted by $a_{k} \subseteq \mathcal{A}$. Note that in Eq.(\ref{eq:outer}), we aim to learn a global model $\mathcal{A}$ in a federated learning setting, which formulates our outer optimization. However, in the inner optimization given in Eq.(\ref{eq:inner}), the objective of each client is to optimize its local model's architecture $a_{k}$ and its associated parameters $v_{k}$ while benefiting from the global model $w^{*}$. 
 
\paragraph{Definition of Supernet $\mathcal{A}$ and Child Neural Architecture $a_k$} As a tractable, yet general case study, SPIDER reuses the DARTS architecture space as Supernet $\mathcal{A}$: there are 8 cells, and each cell consists of multiple edges; each edge connects two intermediate representations (node) by a mixture of multiple operations frequently used in various modern CNNs (e.g., sep convolution 3x3, sep convolution 5x5, dil convolution 3x3, skip connection, max pool 3x3, avg pool 3x3); the mixture uses \text{softmax} over all possible operations to relax the categorical discrete candidate to a continuous search space. More precisely, $\mathcal{A}$ contains a set of edges ${\{e_1,...,e_E\}}$, and each edge $e$ has multiple operations ${\{o_1,...,o_O\}}$. Based on this definition, $a_k$ maintains the operation-level granularity: $a_k$'s edge set space is a subset of $\mathcal{A}$'s edge set space, and the operation set in $a_k$'s each edge may also be a subset space.


\paragraph{The difficulty of jointly optimizing the architecture $a_k$ and related weight parameters $v_k$} 
The key difference of our formulation from existing bi-level optimization for FL (e.g., \cite{li2021ditto}) is that in our case, $a_{k}$ is also a learnable parameter (Eq.(\ref{eq:inner})). We assume each client can have an evolving architecture $a_{k}$, i.e., Eq.(\ref{eq:inner}) has to optimize the architecture $a_{k}$ and its related weight parameters $v_{k}$ jointly, while using complete Supernet-based global model weights, $w$. SPIDER addresses this challenge by the orchestration of SPIDER-Trainer and SPIDER-Searcher.

\subsection{SPIDER Trainer: Federated Training on Heterogeneous Architectures}
\label{SPIDER_Train}
In this section, we describe SPIDER trainer, an architecture-personalized training framework that can collaboratively train heterogeneous neural architectures across clients.

To clearly show how SPIDER handles the optimization difficulty of Eq.(\ref{eq:inner}), we first downgrade the objective to the case that all clients use \textit{predefined} (fixed) heterogeneous architectures (derived from the Supernet $\mathcal{A}$). More specially, we reduce the aforementioned optimization framework in Eq.(\ref{eq:inner}) and Eq.(\ref{eq:outer}) to the following:
\begin{align}
\label{eq:inner1}\min_{v_{k}}&\quad h_{k}(v_{k}, a_k; w^{*}, \mathcal{A}) = F_{k}(v_{k}) + \frac{\lambda}{2} || v_{k} - w_{share}^{*}||^2 
\\\label{eq:outer1}\text{s.t.}&\quad w^{*}\in\arg\min_{w}G\left(F_1(w, \mathcal{A}),. . .,F_{K}(w, \mathcal{A})\right),
\end{align}
where local model's weights $v_{k}$ are regularized towards the global model $w^{*}_{share}$, where $w^{*}_{share} = w^{*} \odot a_{k}$ (i.e., only using the weight parameters of the operation set space overlapping (sharing) with  $a_k$). Also, $\lambda$ is the regularization hyper-parameter. Note that, now, only $v_k$ needs to be optimized in Eq.\ref{eq:inner1}, while $a_k$ is fixed during the optimization. 

We then solve Eq.\ref{eq:inner1} and Eq.\ref{eq:outer1} alternately. We summarize this optimization procedure as SPIDER-Trainer with a detailed pseudo code illustrated in Algorithm \ref{alg:SPIDER}. In this algorithm, we can note that the global model (line \#12) and the local model (line \#14) are updated alternately. The strength of this algorithm lies in its elaborate design, which provides the following key benefits:

\begin{algorithm}[htb]
    \caption{\textbf{SPIDER Trainer}}
    \begin{small}
        \begin{algorithmic}[1]
            \STATE \textbf{Initialization:} initialize $K$ clients with the $k$-th client has a global model $w_k$ using Supernet $\mathcal{A}$, and a local model ${v_k}$ using subnet $a_k$ (set $a_k = \mathcal{A}$ at the begining); $E$ is the number of local epochs; $T$ is the number of rounds; $T_s$ number of rounds to start search; $\tau$ is the recovery periods in the units of rounds.
            \STATE \textbf{Server executes:}
            \begin{ALC@g}
            \FOR{each round $t = 0, 1, 2, ..., T-1$}
            \FOR{each client $k$ \textbf{in parallel}}
            \STATE $w_{k}^{t+1}  \leftarrow \text{ClientLocalSearch}(k, w^t, t)$
            \ENDFOR
            \STATE $w^{t+1} \leftarrow \sum_{k=1}^{K} \frac{N_{k}}{N} w_{k}^{t+1}$
            \ENDFOR
            \end{ALC@g}
            \STATE 
            \STATE \textit{function} \textbf{ClientLocalSearch}($k$, $w^t$, $t$): // \textit{Run on client $k$}
            \begin{ALC@g}
            \FOR{$e$ in epoch}
            \FOR{minibatch in training and validation data}
                \STATE $a_{k}^{t} = \textbf{ProgressiveNAS}(a_{k}^{t}, T_s, \tau, t)$
                \STATE Update Global model: $w^{t+1} = w^t - \eta_w \nabla_w \mmL_{k}^{\mathrm{tr}}(w^t)$
                \STATE $w^{t+1}_{share} = w^{t+1} \odot a_{k}^{t} \text{ //weight sharing} $
                \STATE Update Local Model: $\text{          } v_{k}^{t+1} = v_{k}^{t} - \eta_v
                \left(\nabla_v\mmL_{k}^{\mathrm{tr}}(v_{k}^{t}) + \lambda (v_{k}^{t} - w^{t+1}_{share}) \right)$ 
            \ENDFOR
            \ENDFOR
            \STATE \textbf{return} $w$ to server
            \end{ALC@g}
        \end{algorithmic}
    \end{small}
\label{alg:SPIDER}
\end{algorithm}



\paragraph{(1) Enabling regularization between an arbitrary personalized architecture and the global model} Most importantly, SPIDER-Trainer connects each personalized model with the global model by enabling the regularization between two different architectures: an arbitrary personalized architecture for the local model $a_{k}$ of client $k$ and the global model with Supernet $\mathcal{A}$. This is done by weight sharing. More specially, in Eq.\ref{eq:outer1}, $w^{*}_{share} = w^{*} \odot a_{k}$, which provides us the global model's weight parameters for the connections/edges common between the child neural architecture $a_k$ and the global model's Supernet $\mathcal{A}$. $w^{*}_{share}$ is essentially used to regularize a subnet ($a_k$) model parameters $v_k$ towards the global model shared/common parameters $w^{*}_{share}$, as shown in Eq. \ref{eq:outer1}.

\paragraph{(2) Avoiding heterogeneous aggregation} SPIDER-Trainer avoids the aggregation of heterogeneous model architectures at the server side. As such, no sophisticated and unstable aggregation methods are required (e.g., masking, knowledge distillation \cite{Lin2020EnsembleDF}, etc.), and it is flexible to use other aggregation methods beyond FedAvg (e.g., \cite{Karimireddy2020SCAFFOLDSC,Reddi2021AdaptiveFO}) to update the global model. 

\paragraph{(3) Enabling architecture privacy} In this algorithm, only the global model is transmitted between the client and the server. This enables architecture privacy because each client's architecture is hidden from server and other clients. 

\paragraph{(4) Potential robustness to adversarial attacks} The weight sharing-based regularization not only yields the benefit of personalization in FL, but also makes the FL framework more robust to adversarial attacks. Its robustness advantage comes from its ability to keep the local model private and regularizing towards the global model based on its regularization parameter, as shown before by architecture-homogeneous bi-level optimization~\cite{li2021ditto}.


\subsection{SPIDER-Searcher: Personalizing Architecture}
\label{sec_progressive_nas}
Although SPIDER trainer is able to collaboratively train heterogeneous architectures, manual design of the architecture for each client is impractical or suboptimal. As such, we further add a neural architecture search (NAS) component, SPIDER-Searcher, in Algorithm \ref{alg:SPIDER} (line \#11) to adapt $a_k$ to its local data distribution in a progressive manner. We now present the details of SPIDER-Searcher.

\paragraph{Progressive Neural Architecture Search} Essentially, SPIDER-Searcher dynamically changes the architecture of $a_k$  during the entire federated training process. This is feasible because the weight sharing-based regularization can handle an arbitrary personalized architecture (introduced in Section \ref{SPIDER_Train}). Due to this characteristic, SPIDER-Searcher can search $a_k$ in a progressive manner (shown in Figure 1): \textbf{Phase 1}: At the beginning, $a_k$ is set equal to Supernet $\mathcal{A}$. The intention of SPIDER-Searcher in this phase is to warm up the training of the initial $a_k$ so it does not change $a_k$ for a few rounds; \textbf{Phase 2}: After warming up, SPIDER-Searcher performs edge-by-edge search gradually. In each edge search, only the operation with the highest impact to the accuracy is kept. It also uses a few rounds of training as a recovery time before proceeding the next round of search. This process continues until all edges finish searching; \textbf{Phase 3}: After all edges finish searching, SPIDER-Searcher does not change  $a_k$. This serves as a final training of the searched architecture-personalized model. This three-phase procedure is summarized as Algorithm \ref{alg:algorithm1}. Now, we proceed to elaborate how we calculate the impact of an operation on the Supernet.

\begin{algorithm}[htb]
    \caption{\textbf{SPIDER-Searcher}}
    \begin{small}
        \begin{algorithmic}[1]
            \STATE \textbf{Search Space:} in the architecture $a_k^{t} \subseteq \mathcal{A}$,   $\mathcal{E}$ is the super set of all edges ${\{e_1,...,e_E\}}$, $\mathcal{E}_s$ is the remaining subset of edges that have not been searched, and each edge $e$ has multiple operations ${\{o_1,...,o_O\}}$.
            
            \STATE \textit{function} \textbf{ProgressiveNAS}($a_k$, $T_s$, $\tau$, t)
            \begin{ALC@g}
                \IF{$t$ $\geq$ $T_s$ and $t$ \% $\tau$ == 0 and LEN($\mathcal{E}_s$) $>$ $0$}

                    \STATE $e_i$ = RANDOM ($\mathcal{E}$) \text{ // random selection}
                    \STATE \text{// searching without training}
                    \FORALL{operation $o_j$ on edge $e_i$ }
                        \STATE evaluate validation accuracy of $a_k^t$ when $o_j$ is removed$(ACC_{\setminus o})$
                    \ENDFOR
            \STATE in $e_i$, keep only one operation corresponding to the lowest value of $(ACC_{\setminus o})$, i.e., highest impact.
            \STATE remove $e_i$ from $\mathcal{E}$
            \ELSE
                \STATE return $a_k^{t}$ directly
            \ENDIF
            \STATE \textbf{return} updated $a_k^{t}$ after selection
            \end{ALC@g}

        \end{algorithmic}
    \end{small}
\label{alg:algorithm1}
\end{algorithm}

\paragraph{Operation-level perturbation-based selection} In phase 2, we specify selecting the operation with the highest impact using operation-level perturbation. More specially, instead of optimizing the mixed operation architecture parameters $\alpha$ using another bi-level optimization as DARTS (a.k.a. gradient-based search) to pick optimal operation according to magnitude of $\alpha$ parameters (\textit{magnitude-based selection}), we fix a uniform distribution for $\alpha$ and use \textit{the impact of an operation on the local validation accuracy} (perturbation) as a criterion to search on the edge. This simplified method is much more efficient given that it only requires evaluation-based search rather than training-based search (optimizing $\alpha$). In addition, it avoids inserting another bi-level optimization for NAS inside a bi-level optimization for FL, making the framework stable and easy to tune. Finally, this method avoids suboptimal architecture \cite{wang2021rethinking} lead by magnitude-based selection in differentiable NAS.

%% file: sec/4_results.tex
\section{Experiments}

This section presents the experimental results of the proposed method, SPIDER. All our experiments are based on a non-IID data distribution among FL clients. We have used latent Dirichlet Distribution (LDA), which is a common data distribution in FL to generate non-IID data across clients \cite{he2020fedml}, \cite{yurochkin2019bayesian}. 

\subsection{Experimental Setup}

\paragraph{Implementation and Deployment} We implement the proposed method for distributed computing with nine nodes, each equipped with GPUs. We set this as a  cross-silo FL setting with one node representing the server and eight nodes representing the clients. These clients nodes can represent real-world organizations such as hospitals and clinics that aim to collaboratively search for personalized architectures for local benefits in a privacy-preserving FL manner.

\begin{figure}[h!]
\begin{subfigure}{.48\linewidth}
  \centering
    \includegraphics[width=\linewidth]{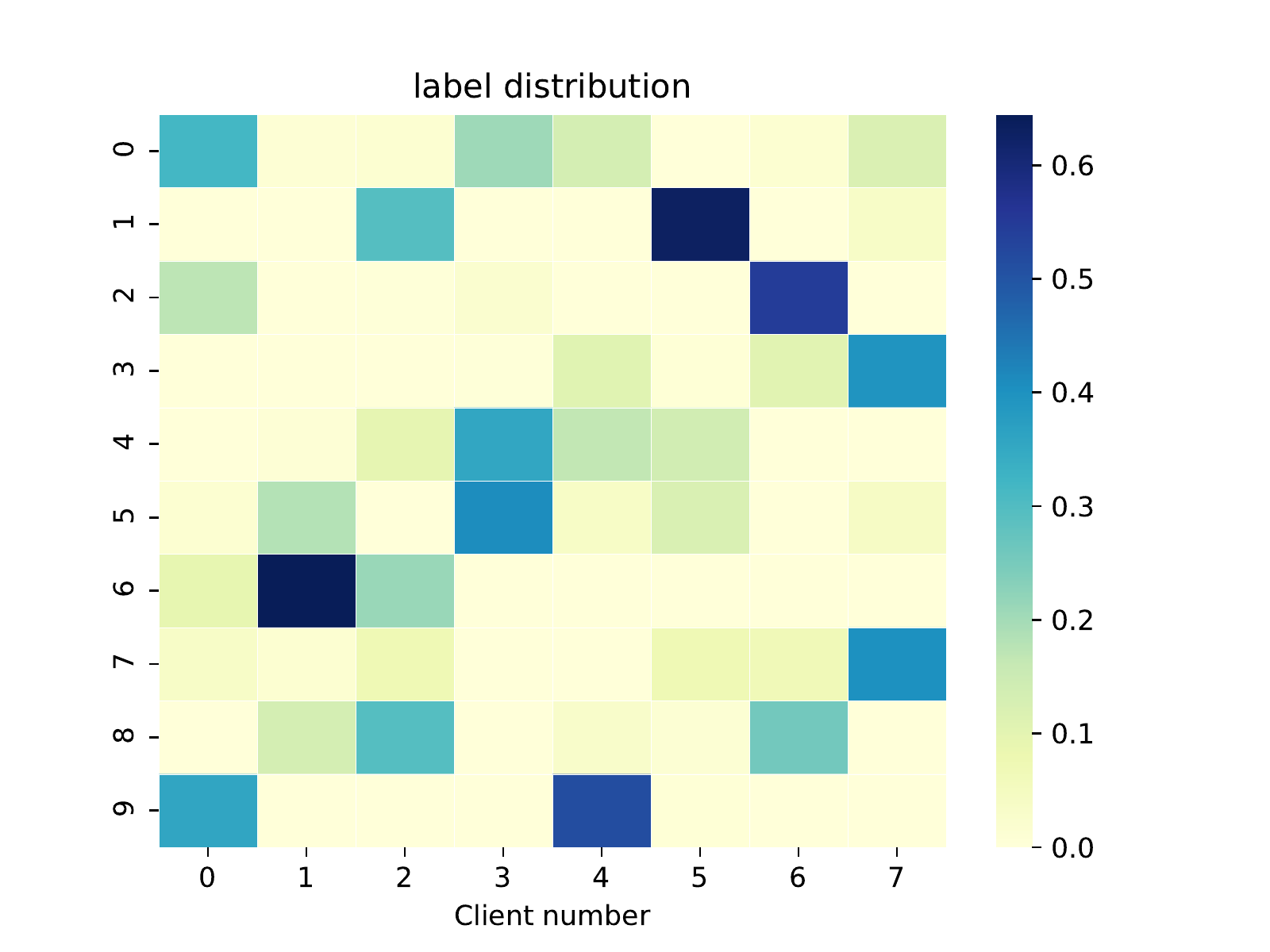}
    \caption{Label distribution per client}
    \label{fig:label_distribution}
\end{subfigure}
\begin{subfigure}{.48\linewidth}
  \centering
    \includegraphics[width=\linewidth]{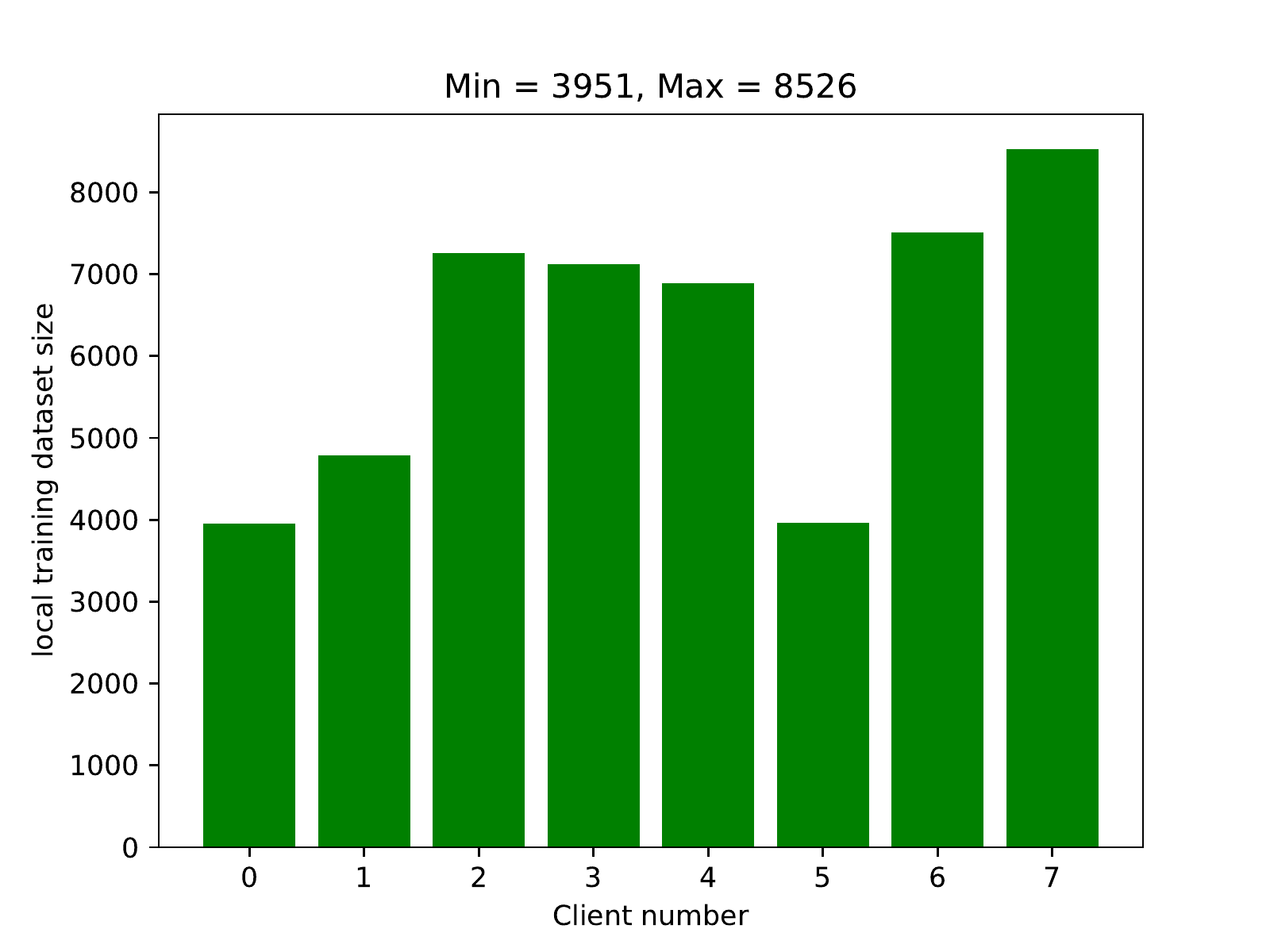}
    \caption{Image distribution per client}
    \label{fig:image_distribution}
\end{subfigure}
\caption{CIFAR10: LDA distribution}
\label{fig:figure1}
\vspace{-0.5cm}
\end{figure}

\begin{figure*}
\begin{subfigure}{.50\linewidth}
  \centering
    \includegraphics[width=0.80\columnwidth]{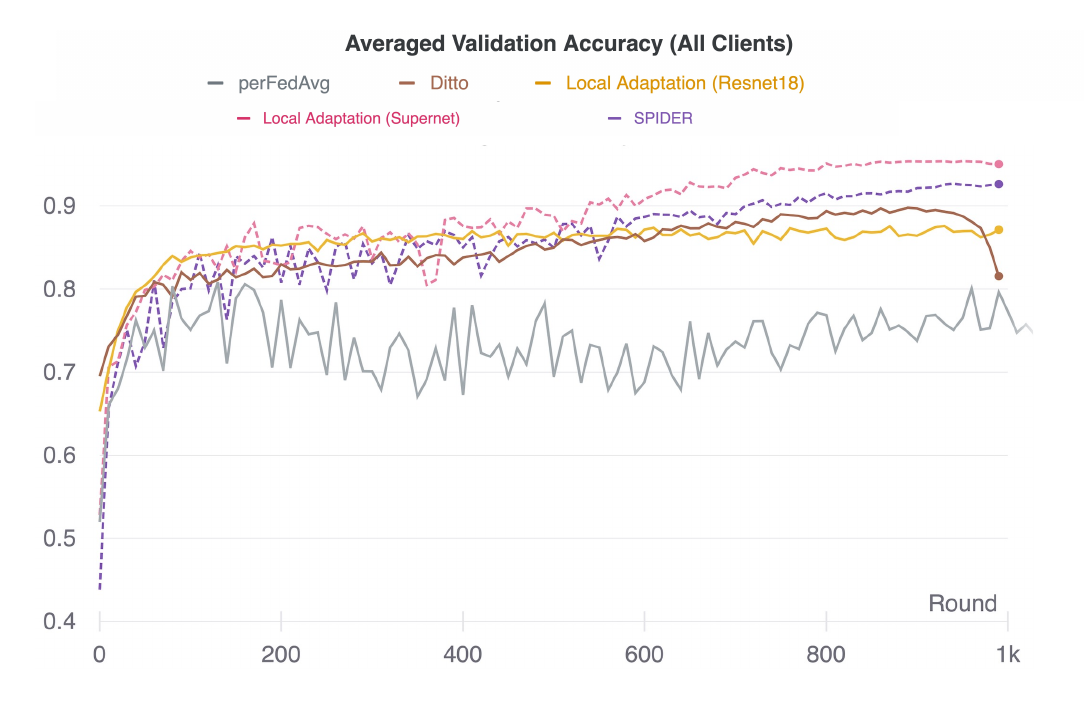}
    \caption{Average Validation Accuracy comparison}
\end{subfigure}
\begin{subfigure}{.50\linewidth}
  \centering
    \includegraphics[width=0.80\columnwidth]{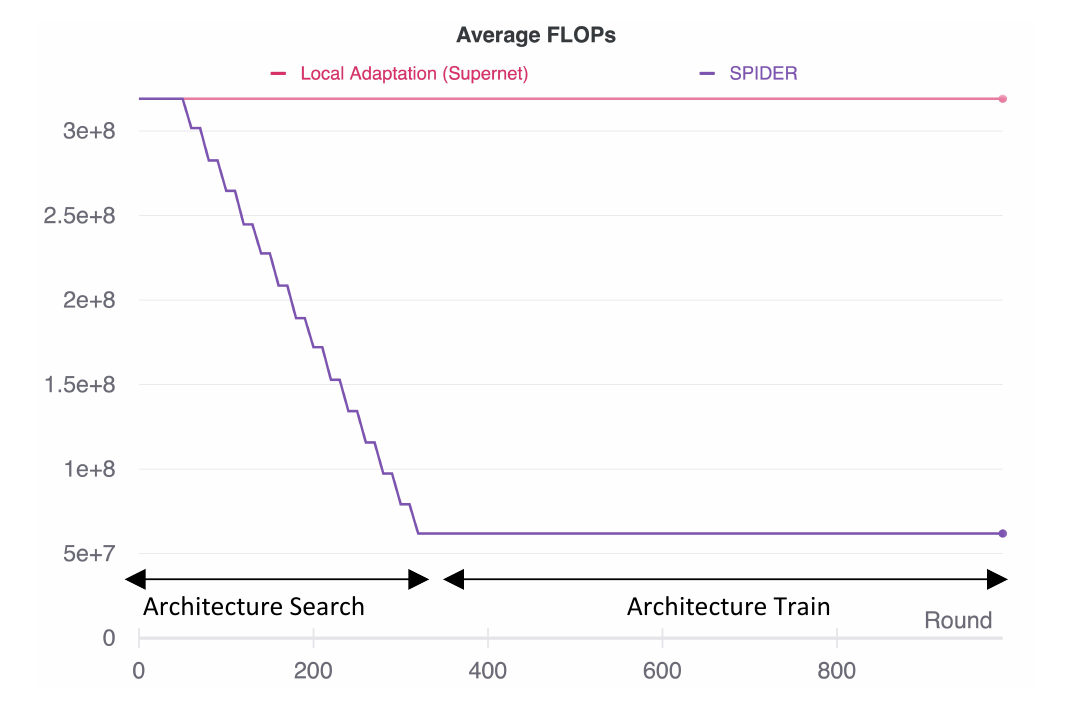}
    \caption{Architecture search vs. Architecture train Phase}
\end{subfigure}
\label{fig:comparison}
\caption{Comparison of our proposed method, SPIDER, with other state-of-the-art personalization methods (perFedAvg, Ditto, and local adaptation). The left figure shows the average validation accuracy comparison between our proposed method SPIDER and the other state-of-the-art methods; perFedAvg, Ditto, and local adaptation.
The right figure illustrates the architecture search (progressive perturbation of the Supernet) and derived child model's architecture train phase of the proposed method.}
\label{fig:figure1}
\vspace{-0.5cm}
\label{fig:comparison}
\end{figure*}

\begin{figure*}
\vspace{0.3cm}
\begin{subfigure}{.32\linewidth}
  \centering
    \includegraphics[width=.98\columnwidth]{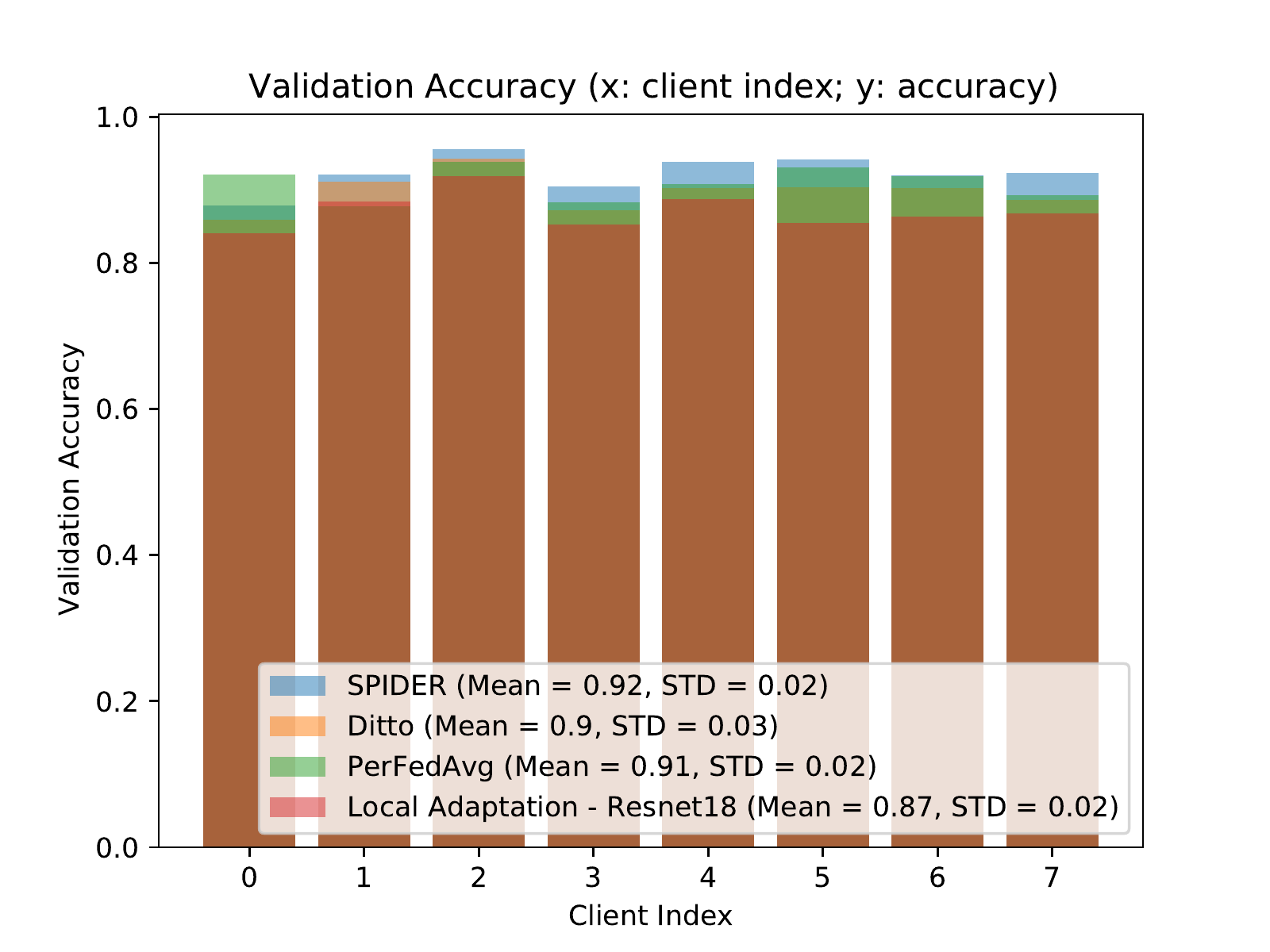}
    \caption{Average Validation Accuracy per client}
\end{subfigure}
\begin{subfigure}{.32\linewidth}
  \centering
    \includegraphics[width=.98\columnwidth]{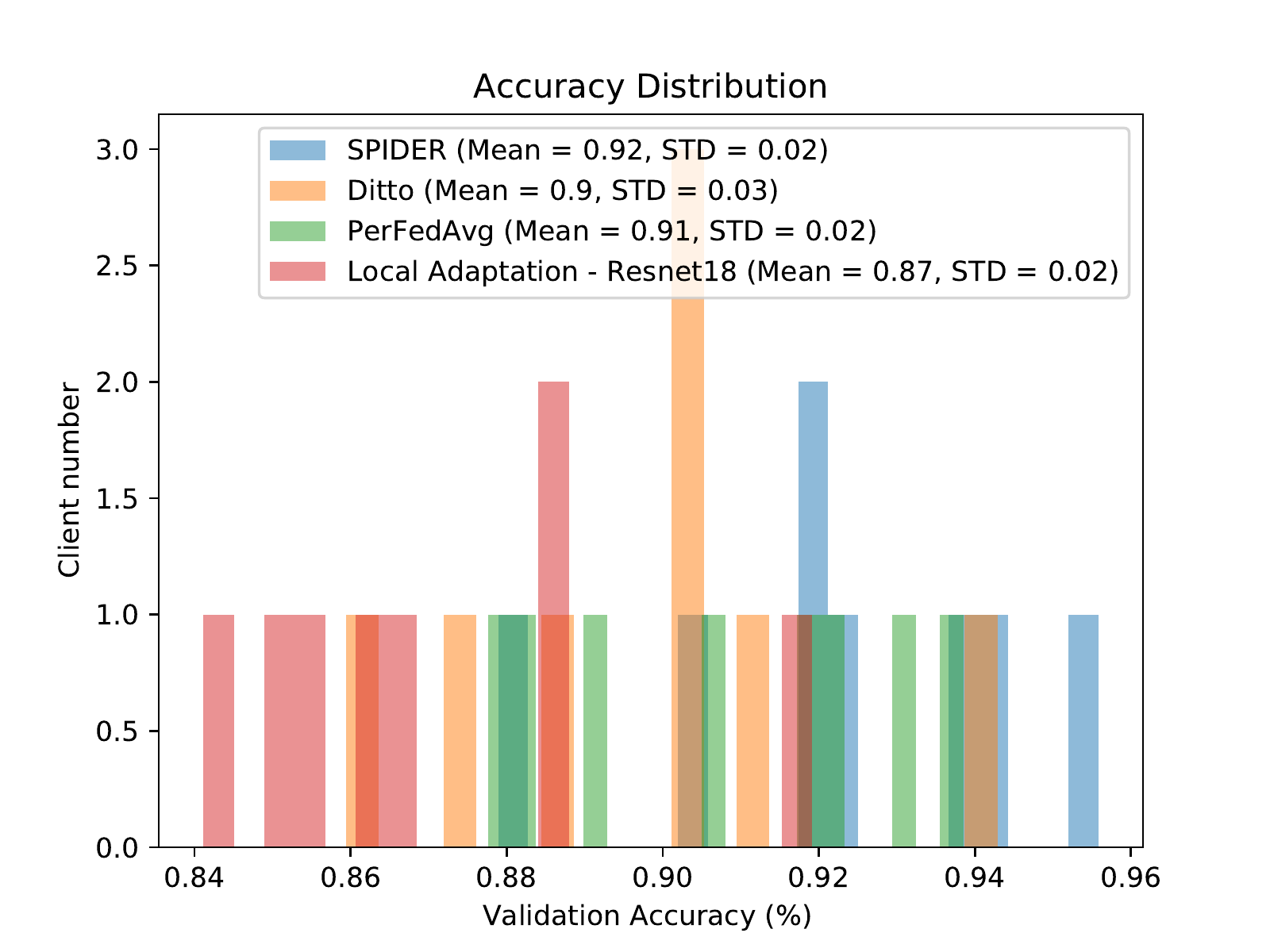}
    \caption{Average Validation Accuracy distribution}
\end{subfigure}
\begin{subfigure}{.32\linewidth}
  \centering
    \includegraphics[width=.98\columnwidth]{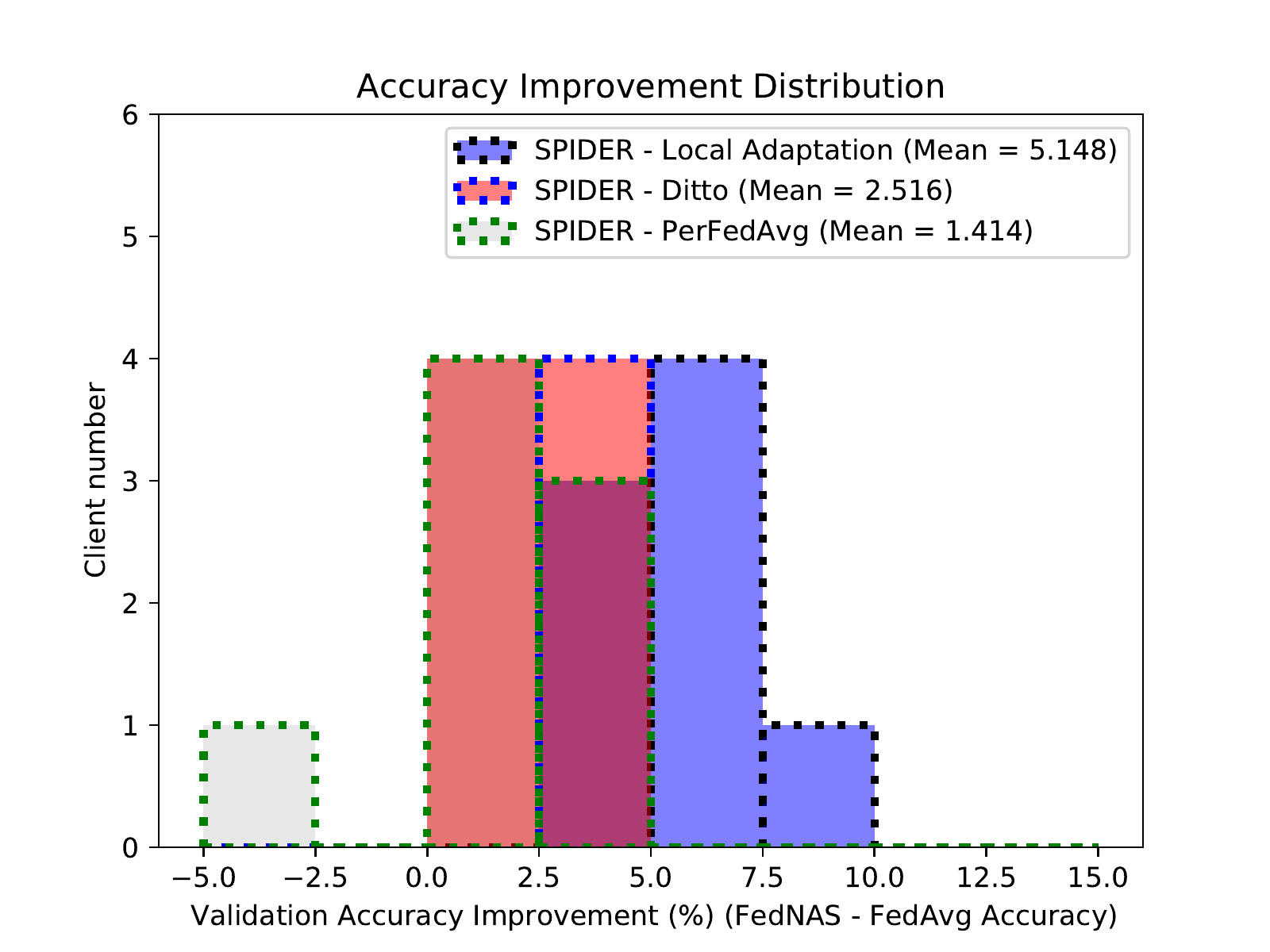}
    \caption{Accuracy improvement of SPIDER}
\end{subfigure}
\vspace{-0.2cm}
\caption{
%
Validation Accuracy Analysis  
}

\label{fig:vald_acc_analysis}
\end{figure*}


\paragraph{Task and Dataset} We perform an image classification task on the CIFAR10 dataset that consists of 60000 32x32 color images in 10 classes, with 6000 images per class. We generate non-IID data across clients by exploiting LDA distribution with parameter $(\alpha = 0.2)$  for the training data of CIFAR10. The actual data distribution has been shown in figure ~\ref{fig:figure1}. Sub-figure ~\ref{fig:label_distribution} represents the label distribution across clients, where a darker color indicates more images of that class/label. The other sub-figure ~\ref{fig:image_distribution} represents the total number of data samples preset at each client. We explore other datasets in the Appendix~\ref{cifar100-dataset}.

For personalized architecture experiments with SPIDER, we split the total training data samples present at each client into training (50\%), validation (30\%), and testing sets (20\%). For other personalization schemes used for comparison, we do not need validation data. Therefore, we split the data samples of each client with training (80\%) and test (20\%) for a fair comparison. In addition, we fix the non-IID dataset distribution in all experiments for a fair comparison. We provide Hyperparameter search details in Appendix ~\ref{HP}.


\begin{table*}[h!]
\centering
    \caption{Average local validation Accuracy Comparison of Personalized Federated NAS with other personalization techniques}
\resizebox{.75\textwidth}{!}{
    \centering
    \begin{tabular}{lcccc}
    \toprule
    \textbf{Method} & \textbf{Average Accuracy} & \textbf{Parameter Size} & \textbf{FLOPs} & \textbf{Estimated Model Size}\\
     \midrule
        {Local Adapation - Supernet}  & 0.95±0.01 & 1.9M  & 319M & 104MB\\
       \cmidrule{1-5}
       {SPIDER}  & \textbf{0.92±0.02} & \textbf{345K}  & \textbf{62M} & \textbf{14MB} \\
       \cmidrule{1-5}
       Local Adaptation - ResNet18 & 0.87±0.02 & 11M  & 76M &  44MB\\ 
       \cmidrule{1-5}
       Ditto - ResNet18 & 0.90±0.03 & 11M  & 76M & 44MB \\
       \cmidrule{1-5}
       perFedAvg - ResNet18 & 0.91±0.02 & 11M  & 76M &  44MB \\
      \bottomrule
    \end{tabular}
}
    \label{tab:accuracy comparison}
\vspace{-0.3cm}
\end{table*}

\subsection{Results on Average Validation Accuracy}

Here, we report the comparison of our proposed method SPIDER with the other state-of-the-art personalized methods; Ditto, perFedAvg, and local adaptation. Since these schemes use a pre-defined architecture, we use the Reset18 model because of its comparable model size. We also explore local adaptation with the complete Supernet. This exploration will help us investigate how much performance drop we get if we use a smaller but personalized model compared with a locally adapted complete Supernet. 

\textbf{Average Accuracy} In Figure ~\ref{fig:comparison}, we report average of the validation accuracy calculated on client's test dataset using the personalized architectures (architecture visualizations provided in Appendix~\ref{Arch}). The right sub-figure in Fig. ~\ref{fig:comparison} illustrates the comparison of the proposed method with the state-of-art methods; Ditto, local adaptation, and perFedAvg. We note that local adaptation with Supernet provides the highest average accuracy. We expect it from the Supernet because of its 7 operations-based mixed operation formulations. This result helps us investigate the performance reduction with our proposed method. We observe that our proposed method shows a performance reduction of 2.7\% in average validation accuracy with the benefit of a reduced average number of FLOPs and model size. The models we use during training are much smaller and require less memory as well as computation.

From empirical results, we observe that the proposed approach of architecture personalization outperforms the other state-of-the-art personalization methods; Ditto, perFedAvg, and local adaptation. Among these personalization schemes, perFedAvg yields the highest performance. Although in Figure~\ref{fig:comparison} the performance curve is lower, it gains 90.9\% accuracy around 1500 rounds. We have added its accuracy plot for 1500 rounds in Appendix \ref{perFedAVg_graph}. Compared with perFedAvg, we obtain 1.4\% higher average accuracy. Moreover, Ditto and local adaptation yield a 90\% and 87\% average accuracy, respectively.

We also analyze the performance of these methods on a client basis. We plot validation accuracy of each client, accuracy distribution of each method, and accuracy improvement of the proposed method SPIDER with the other personalization methods in Figure ~\ref{fig:vald_acc_analysis}. In the first sub-figure, we observe that out of 8 total clients, our method yields higher accuracy for 7 clients. For the other one clients (client number 0), perFedAvg outperforms our method, respectively. In the second sub-figure, we plot accuracy distribution. The motivation behind this plot is to observe the standard deviation of these methods and their concentration of accuracy values. We note that SPIDER dominates the right side, higher accuracy region, and it has almost the same spread as perFedAvg and local adaptation baselines. 

The third sub-figure illustrates the accuracy improvement, subtraction of the proposed method's accuracy values from the other personalized methods on a client-basis. In comparison to perFedAvg, for 4 client, the accuracy improvement with the proposed method is within 0-2.5\% region, and for 3 clients, it is between 2.5-5\% regions. There is only one client for which perFedAvg outperforms our method in the region of 2.5-5\%. For the rest of the methods, our method yields higher performance for all the clients.

For personalization, in addition to average accuracy, the standard deviation (std) is considered an important metric. Therefore, we also report the standard deviation for each method in Table ~\ref{tab:accuracy comparison}. Besides the local adaptation with Supernet that yields 0.01 standard deviation, our method yields 0.02 std, which is the same for the other two personalization baselines; PerFedAvg and local adaptation (resnet18). However, Ditto has the highest deviation (0.03) among clients. Although we achieve the same standard deviation as other baselines but with the benefit of providing higher average validation accuracy.

\subsection{Results on Efficiency}

\paragraph{Average FLOPs} In Figure ~\ref{fig:comparison}, we report average flops to gauge the architecture search and train phase. Following Algorithm ~\ref{alg:SPIDER}, Phase 1 continues for 60 rounds, Phase 2 for another 260 rounds, Phase 3 for the next 680 rounds. We used a recovery time of around 20 rounds. In centralized perturbation-based NAS \cite{wang2021rethinking}, it was empirically found that 5 epochs are sufficient for fine-tuning the Supernet. However, from empirical results in FL, we found 20-30 communication rounds to be a reasonable number of rounds for recovery. The Average FLOPs vs. number of rounds figure illustrates the reduction in the size of computations as the search proceeds. The reduction in the average size of computations has been from 319M to 62M. However, for the local adaptation, the average FLOPs remain the same. Likewise, the average parameter size drops from 1.9M to 345K.  


\paragraph{Model Size} Estimated model size includes the memory size of a forward pass and parameter size of that model. It essentially corresponds to how much memory a trained model with trained weights would occupy to obtain inference. Supernet takes the highest amount of space, 104MB, ResNet18 takes 44MB, and the average model size from our proposed method occupies 14MB space. The reason we report this number is that even if Supernet has less parameter size, 1.9M (due to many non-parameterized operations, i.e., skip connection, max pool, avg pool), it requires a vast number of FLOPs to compute mixed operations on the feature maps of all the operations used in our search space. Therefore, the estimated model size can better represent the actual memory/compute of these models. Since we are working on a cross-silo setting, search cost (two models on each client) may not be a concern since devices in cross-silo are rich in resources. However, the proposed search yields architectures that are smaller in size and inference efficient, which can be beneficial for many business models.

%% file: sec/5_conclusions.tex
\section{Conclusion}
We proposed SPIDER, an algorithmic framework that can search personalized neural architecture for FL. SPIDER specializes a weight-sharing-based global regularization to perform progressive neural architecture search. Experimental results demonstrate that SPIDER outperforms other state-of-the-art personalization methods, and the searched personalized architectures are more inference efficient.

%% file: sec/X_supplementary.tex
\appendix

\setcounter{page}{1}

\twocolumn[
\centering
\Large
\vspace{0.5em}Appendix \\
\vspace{1.0em}
] 
\appendix
\section{Search Space}
\label{search_space}
\begin{figure}[h!]
  \centering
    \includegraphics[width=\linewidth]{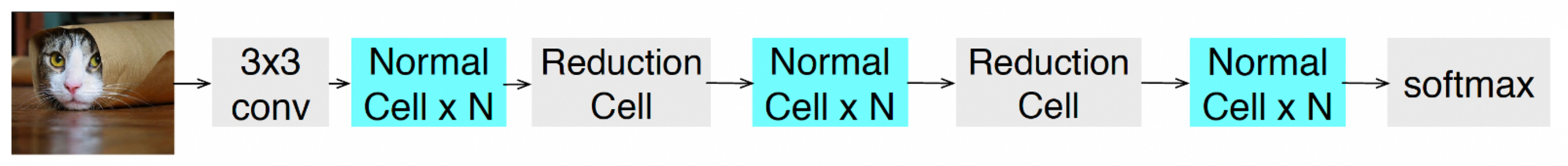}
\caption{Supernet $\mathcal{A}$ search space. Note that since we have 8 cell based search space, N = 2 for our construction.}
\label{fig:system_design}
\vspace{-0.5cm}
\end{figure}

As mentioned in Section~\ref{sec_bilevel}, we have used DARTs search space \cite{liu2018darts} in our proposed work. During the search, we are using a total of 8 cells. In this search space, there are two types of cells, normal cell, and reduction cell. For the Supernet $\mathcal{A}$ construction, there are two normal cells accompanied by one reduction cell and so on, as shown in Figure~\ref{fig:system_design}. In our construction, the reduction cell is used at positions 3 and 6, and at the remaining positions, 1,2,4,5,7,8, normal cell is placed. Furthermore, when we perform perturbation, we perturb both of the cells, normal and reduction cells, one by one at that particular communication round. In addition, the same cells follow the same construction; therefore, if one operation is selected for one normal cell at a particular edge, the same operation will be selected for the same edge at all normal cells.

\section{Hyperparameters}
\label{HP}
\subsection{Hyper-parameters for CIFAR10.} For empirical results of CIFAR10, we use a batch size of 32 for all our experiments. Furthermore, we use a learning rate in the search range of $\{0.1, 0.3, 0.01, 0.01\}$ for SPIDER and local adaptation with Supernet. For SPIDER, we used $\lambda$ search from the set of $\{0.01, 0.1, 1\}$. For the other personalized schemes such as Ditto, perFedAvg, and local adaptation with Resnet18, we searched learning rate over the set $\{0.1, 0.3, 0.01, 0.03, 0.001, 0.003\}$. The reason for having a larger set of learning rates for these methods is that we found $0.001$ and $0.003$ work better for these methods. For SPIDER and local adaptation with Supernet, a learning rate of 0.001 and 0.003 did not seem a better choice in early stop explorations. For Ditto, we used $\lambda$ from the set $\{0.01, 0.1, 1\}$. We used 1000 rounds for the reported results except for perFedAvg, for which we observed convergence for 1400 rounds.

\section{Remaining Results on  CIFAR10} 
\subsection{Personalized Architectures}
\label{Arch}
\begin{figure*}
\begin{center}
\includegraphics[width=2\columnwidth]{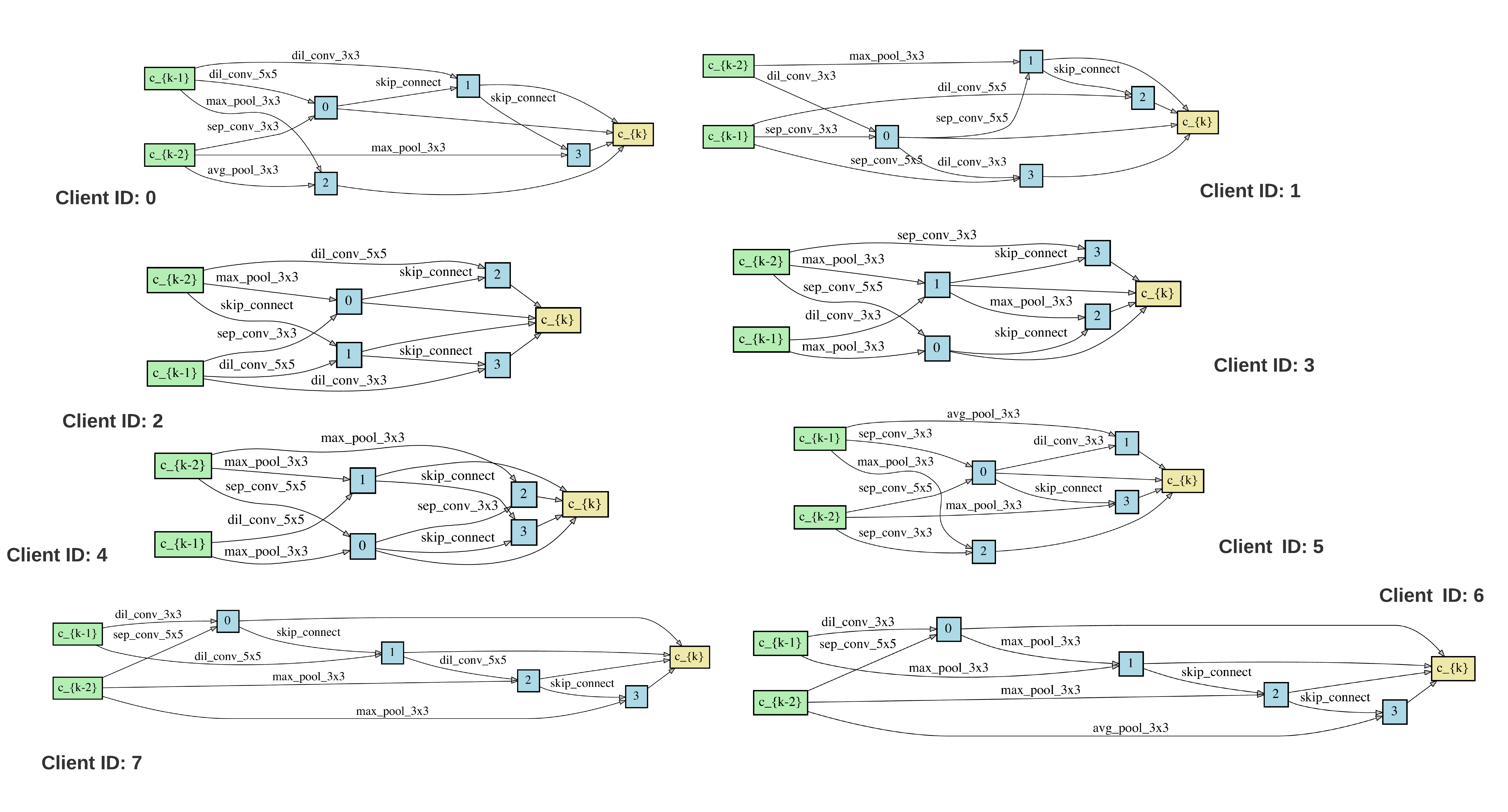}
\end{center}
\caption{
%
Searched Architectures (Normal Cells): Each normal cell $k$ takes the outputs of previous cells, cell $k-2$ and cell $k-1$, as its input. Each cell contains seven nodes: two input nodes, one output node, and four intermediate nodes inside the cell. The output node concatenates all intermediate nodes’ output depth-wise. It can be observed that the searched cells are edge-wise and operation-wise heterogeneous from client to client.}
\label{fig:visualize_architecture}
\end{figure*}

To investigate the architecture heterogeneity, we also visualize the cell connection at each client (Figure \ref{fig:visualize_architecture}). This confirms that the architectures at clients are diverse. This heterogeneity has the potential to bring us the benefit of personalization, each client designs its own architecture to achieve better performance in a statistical non-I.I.D data setting, and model privacy, in a sense that these heterogeneous architectures are not shared with the server.

\begin{figure}[h!]
  \centering
    \includegraphics[width=\linewidth]{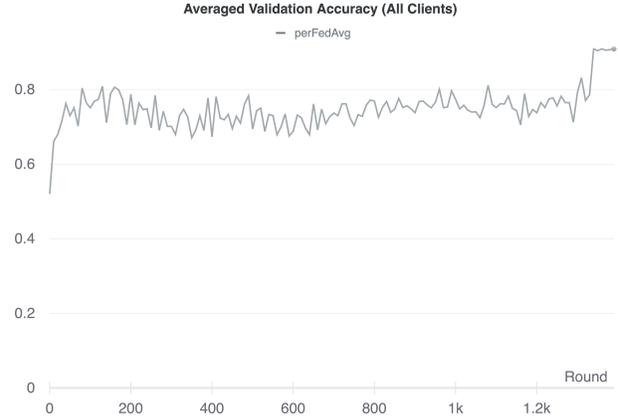}
\caption{perFedAvg validation accuracy versus number of rounds for CIFAR10 dataset.}
\label{fig:perFedAvg}
\vspace{-0.5cm}
\end{figure}

\subsection{perFedAvg results for CIFAR10.}
\label{perFedAVg_graph}
Since perFedAvg took longer to converge, and the results presented in Table~\ref{tab:accuracy comparison} are different from the accuracy value shown in Figure~\ref{fig:comparison}. Therefore we report its accuracy curve in Figure~\ref{fig:perFedAvg} to show its superior performance over Ditto and Local adaptation with resnet18.
\section{Experiments on CIFAR100 Dataset}
\label{cifar100-dataset}
This section presents the experimental results of the proposed method, SPIDER in comparison to local adaptation, perFedAvg and Ditto with CIFAR100 dataset, another dataset explored by researchers for Federated learning \cite{reddi2020adaptive}. All our experiments are based on a non-IID data distribution among FL clients. To generate non-I.I.D data distribution, we have used latent Dirichlet Distribution (LDA). 
\subsection{Experimental Setup}
\paragraph{Implementation and Deployment.} We implement SPIDER for distributed computing with nine nodes, each equipped with a GPU. Similar to CIFAR10 dataset implementation setup, we set up cross-silo FL setting for CIFAR100 dataset with one node representing the server and eight nodes representing the clients. 
\begin{figure*}
\begin{subfigure}{.46\linewidth}
  \centering
    \includegraphics[width=0.80\columnwidth]{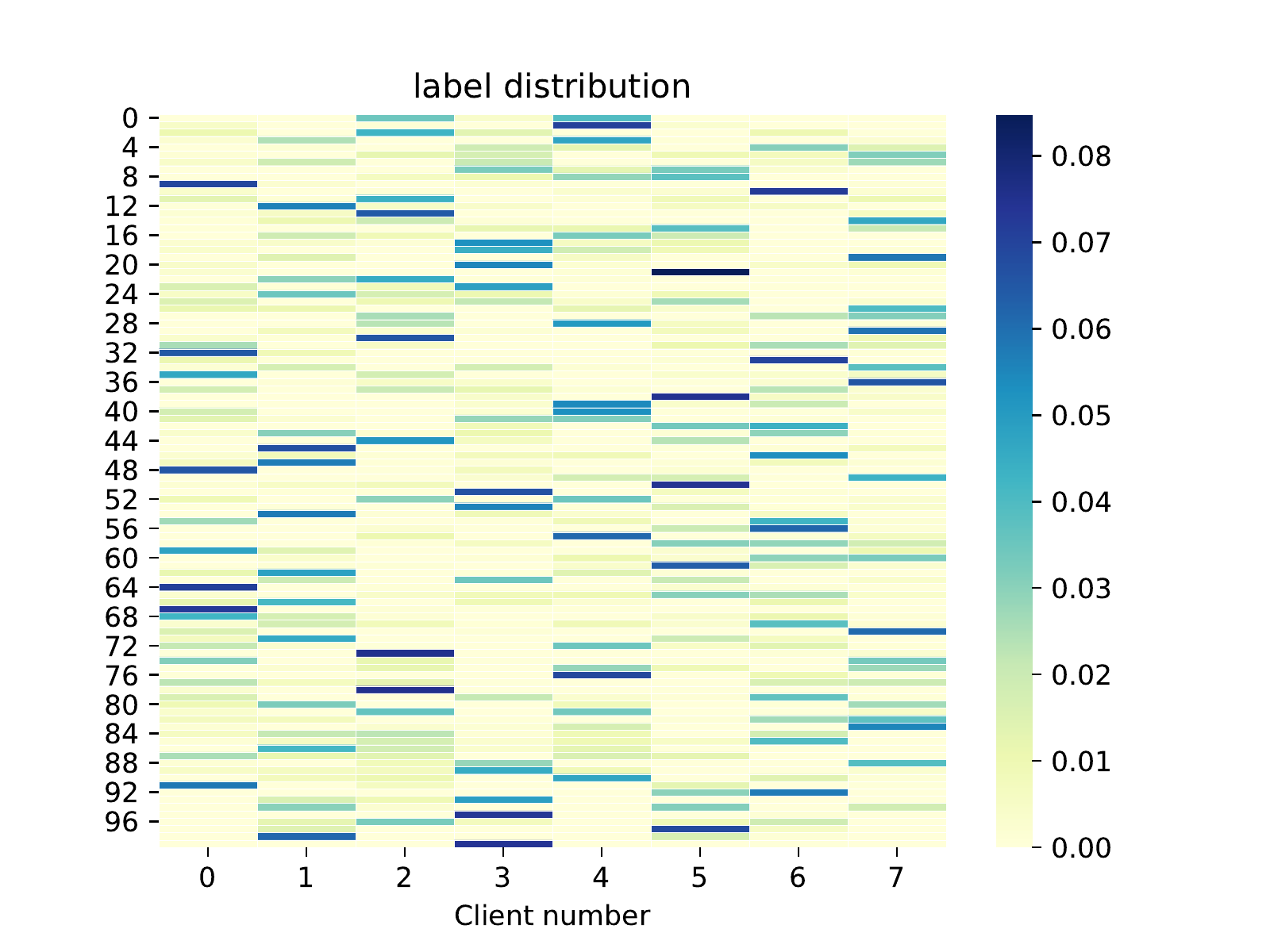}
    \caption{Label distribution per client}
\end{subfigure}
\begin{subfigure}{.46\linewidth}
  \centering
    \includegraphics[width=0.80\columnwidth]{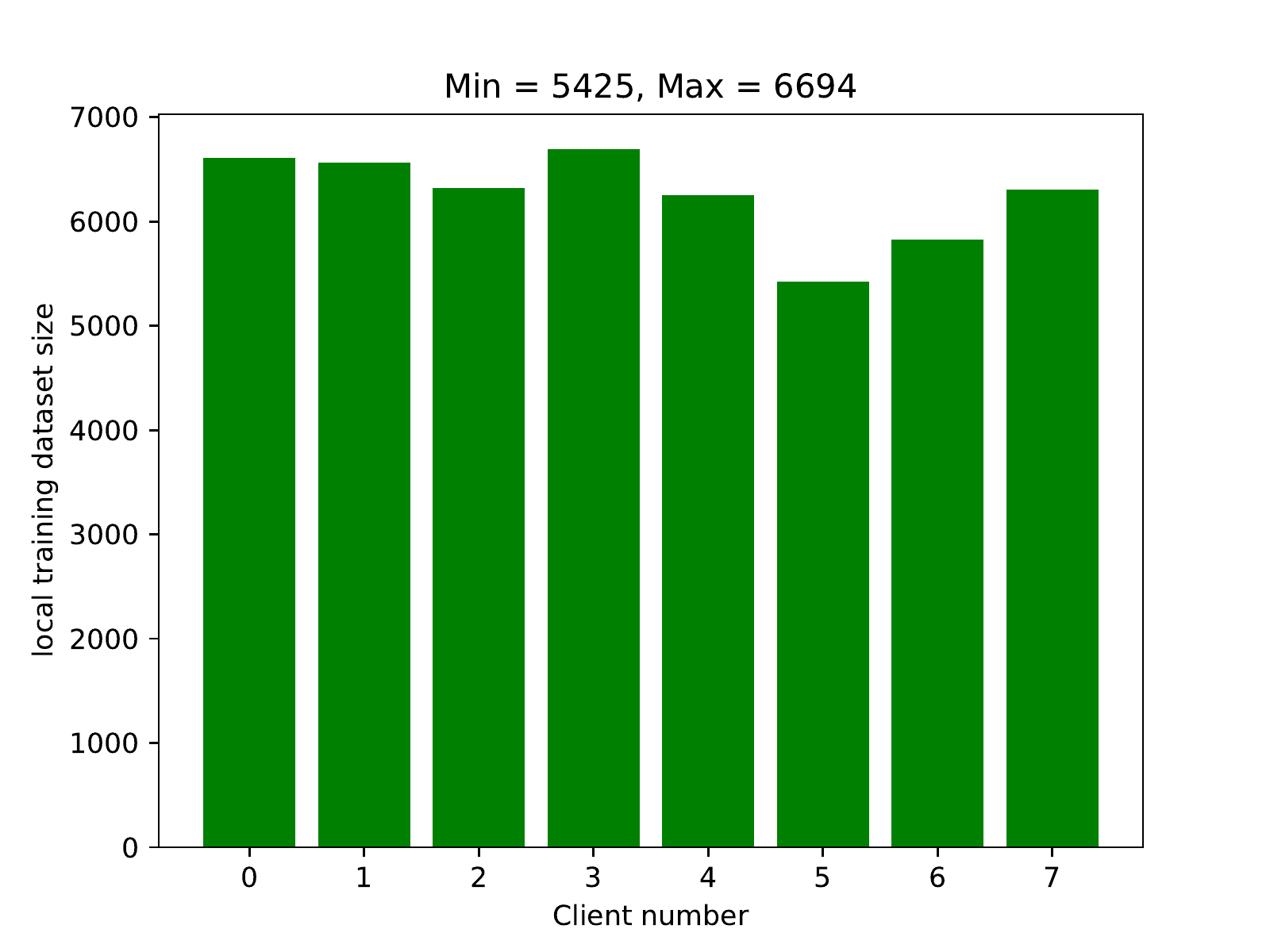}
    \caption{Image distribution per client}
\end{subfigure}
\caption{CIFAR100: LDA distribution}
\label{fig:cifar100}
\vspace{-0.5cm}
\end{figure*}


\paragraph{Task and Dataset}We perform an image classification task on another dataset, the CIFAR100 dataset, that consists of 60000 32x32 color images in 100 classes, with 600 images per class. We generate non-IID data across 8 clients by exploiting LDA distribution with parameter $(\alpha = 0.2)$  for the training data of CIFAR100. The actual data distribution across clients has been shown in figure ~\ref{fig:cifar100}. The right sub-figure represents the label distribution across clients, where a darker color indicates more images of that class/label. The left sub-figure represents the total number of data samples present at each client.


Similar to CIFAR10's experiments setup, for the CIFAR100 dataset, we split the total training data samples present at each client into training (50\%), validation (30\%), and testing sets (20\%). For comparison, we provide results for local adaptation with resnet18, perFedAvg, and Ditto. Since local adaptation, perFedAvg, and Ditto do not require any validation dataset, we split the total training data at each client into (80\%) and testing sets (20\%) for these methods. For a fair comparison, we fix the non-IID dataset distribution in all experiments. Next, we provide Hyperparameter search details for this experiment.

\begin{figure*}
\begin{subfigure}{.46\linewidth}
  \centering
    \includegraphics[width=0.80\columnwidth]{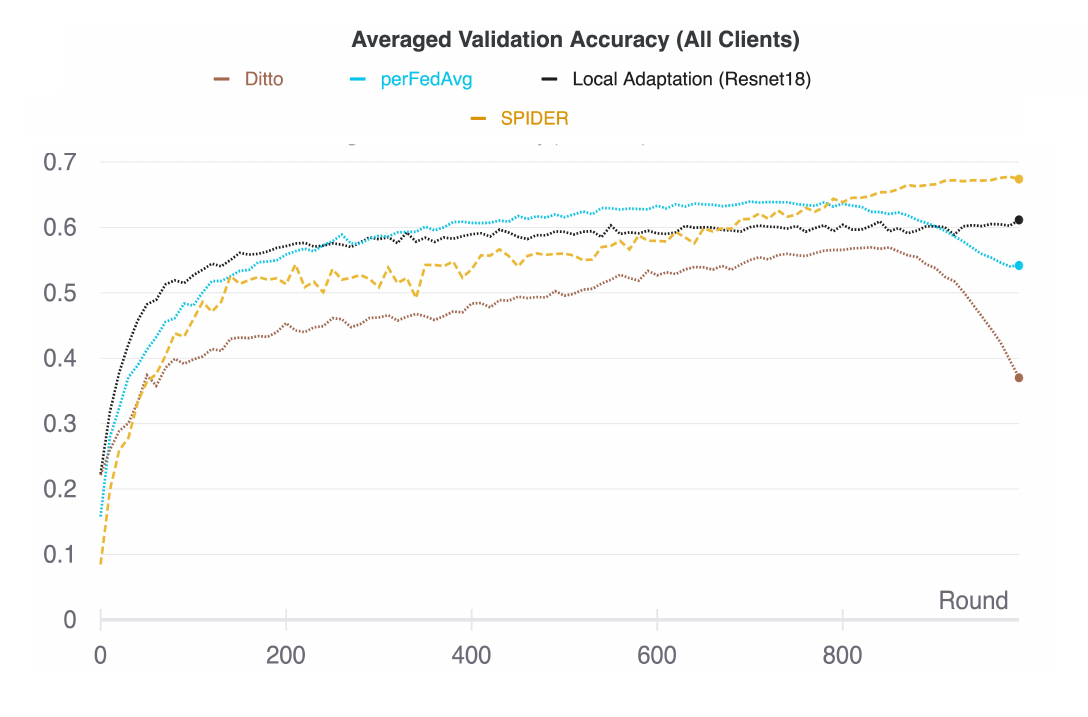}
    \caption{Average Validation Accuracy comparison}
\end{subfigure}
\begin{subfigure}{.46\linewidth}
  \centering
    \includegraphics[width=0.80\columnwidth]{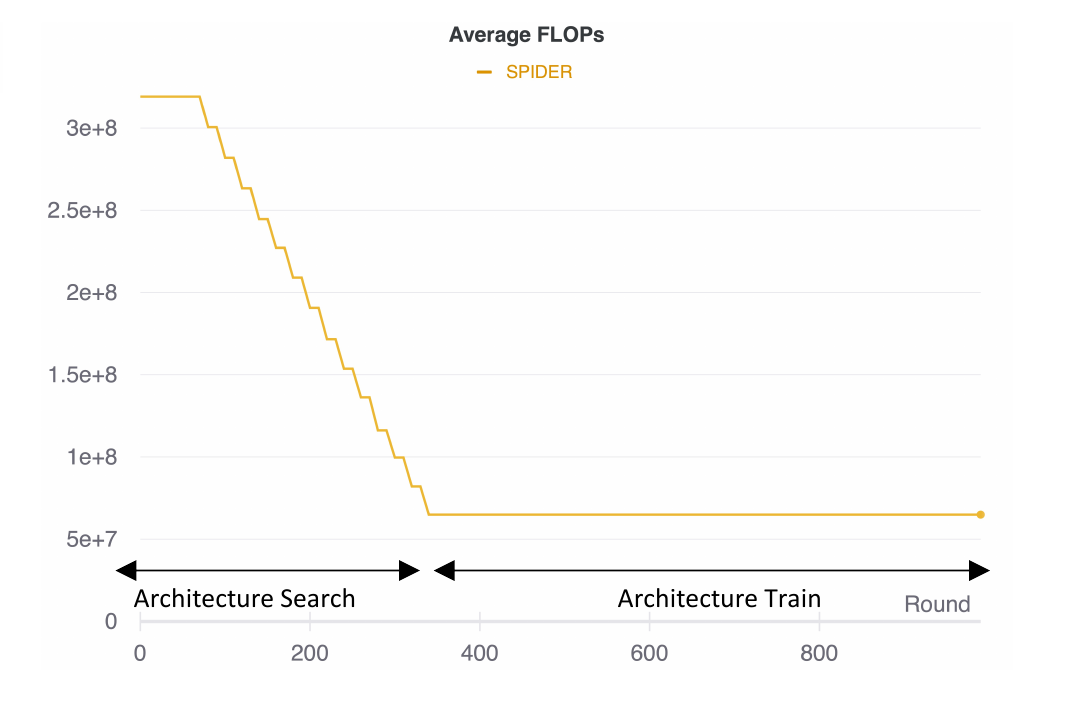}
    \caption{Architecture search vs. Architecture train Phase}
\end{subfigure}
\caption{Comparison of our proposed method, SPIDER, with other state-of-the-art personalization methods (perFedAvg, Ditto, and local adaptation). The left figure shows the average validation accuracy comparison between our proposed method SPIDER and the other state-of-the-art methods; perFedAvg, Ditto, and local Adaptation.
The right figure illustrates the architecture search (progressive perturbation of the Supernet) and derived child model's architecture train phase of the proposed method.}
\label{fig:figure1}
\vspace{-0.5cm}
\label{fig:comparison2}
\end{figure*}

\begin{table*}[h!]
\centering
    \caption{Average local validation Accuracy Comparison of SPIDER with other personalization techniques}
\resizebox{.75\textwidth}{!}{
    \centering
    \begin{tabular}{lcccc}
    \toprule
    \textbf{Method} & \textbf{Average Accuracy} & \textbf{Parameter Size} & \textbf{FLOPs} & \textbf{Estimated Model Size}\\
     \midrule
       {SPIDER}  & \textbf{0.67±0.03} & \textbf{392K}  & \textbf{64M} & \textbf{16MB} \\
       \cmidrule{1-5}
       Local Adaptation - ResNet18 & 0.61±0.03 & 11M  & 76M &  44MB\\
       \cmidrule{1-5}
       Ditto - ResNet18 &0.57±0.03 & 11M  & 76M & 44MB \\
       \cmidrule{1-5}
       perFedAvg - ResNet18 & 0.64±0.03 & 11M  & 76M &  44MB \\
      \bottomrule
    \end{tabular}
}
    \label{tab:accuracy comparison}
\end{table*}



\paragraph{Hyper-parameters} For empirical results of CIFAR100, we use a batch size of 32 for all our experiments. Furthermore, we use a learning rate in the search range of $\{0.1, 0.3, 0.01, 0.01, 0.001, 0.003\}$ for SPIDER. For SPIDER, we used $\lambda$ search from the set of $\{0.01, 0.1, 1\}$. For the local adaptation with Resnet18, perFedAvg and Ditto, we searched learning rate over the set $\{0.1, 0.3, 0.01, 0.03, 0.001, 0.003\}$. For Ditto, we used $\lambda$ from the set $\{0.01, 0.1, 1\}$. We used 1000 number of communication rounds for the reported results.


\subsection{Results on Average Validation Accuracy}
Here, we report the comparison of our proposed method SPIDER with the other state-of-the-art personalized methods; Ditto, perFedAvg, and local adaptation. Given these methods work on a pre-defined architecture, we use the Reset18 model because of its comparable model size to personalized architectures. 

\textbf{Average Validation Accuracy and Personalization.} In Figure ~\ref{fig:comparison2}, we provide the average validation accuracy calculated on each client's local test dataset using the personalized architectures they search with SPIDER. The right sub-figure in Fig. ~\ref{fig:comparison2} demonstrates the comparison of the proposed method with the state-of-art methods; Ditto, local adaptation, and perFedAvg. We note that our method outperforms all three methods by yielding higher accuracy around 67\%.  Similar to our observation with the CIFAR10 dataset, perFedAvg yields the second highest accuracy ~64\%. We also note that contrary to CIFAR10 results, Ditto provided the lowest performance among the three personalization methods. From a personalization perspective, we observe that our method achieves the same standard deviation, 0.032, as perFedAvg. However, we achieve this standard deviation with a 3.5\% higher mean validation accuracy value. The local adaptation and Ditto also yield same standard deviation, which translates to same personalization (fairness) across clients. In a nutshell, experiments with CIFAR10 and CIFAR100 show that SPIDER with architecture personalization has the potential to provide higher performance in terms of higher mean validation accuracy and low/comparable standard deviation (fairness).

Overall, we observed dominance of SPIDER with higher average accuracy over other baselines with CIFAR100 as we did for CIFAR10. However, improvement gap is more prominent with CIFAR100 (3.5\%) as compared to CIFAR10 (1.4\%) with the second dominant personalization method, perFedAvg. Moreover, we also note that with CIFAR100 all baselines yield same standard deviation (0.03) as other personalization baselines, however, in CIFAR10, Ditto had slightly higher standard deviation.




